\documentclass[a4paper,10pt]{article}
\usepackage[T1]{fontenc}
\usepackage[latin1]{inputenc}
\usepackage[nooneline,bf]{caption} 
\usepackage{times}
\usepackage{multicol}
\usepackage{amsmath}
\usepackage{amssymb}
\usepackage[dvips]{graphicx}
\usepackage{epic}
\usepackage{eepic}
\usepackage{epsfig}
\usepackage{xspace}  
\usepackage{color}
\usepackage{mdwlist}
\usepackage{enumitem}

\usepackage{cite}      
\usepackage{url}       
\usepackage{stfloats}  
\usepackage{amsmath}   
\usepackage{array}
\usepackage{algorithm}
\usepackage{algpseudocode}
\usepackage{epstopdf}
\usepackage{booktabs}
\usepackage{multirow}
\usepackage{units}
\usepackage{subfig}

\newenvironment*{mytitle}{\begin{LARGE}\bf}{\end{LARGE}\\[1.5ex]}%
\newenvironment*{myabstract}{\begin{Large}\bf}{\end{Large}\\[2.5ex]}%
\makeatletter
{}

\makeatother

\begin{document}

\begin{mytitle} Few common failure cases in mobile robots \end{mytitle}

Ramviyas Parasuraman \footnote{The author was affiliated with European Organization for Nuclear Research (CERN), Switzerland during this work. Contact: ramviyas@kth.se}, CVAP, Royal Institute of Technology (KTH), Stockholm, Sweden.

\vspace{5ex} 

\begin{myabstract} Abstract \end{myabstract}
A mobile robot deployed for remote inspection, surveying or rescue missions can fail due to various possibilities and can be hardware or software related. These failure scenarios necessitate manual recovery (self-rescue) of the robot from the environment. It would bring unforeseen challenges to recover the mobile robot if the environment where it was deployed had hazardous or harmful conditions (e.g. ionizing radiations). While it is not fully possible to predict all the failures in the robot, failures can be reduced by employing certain design/usage considerations. Few example failure cases based on real experiences are presented in this short article along with generic suggestions on overcoming the illustrated failure situations.

\begin{multicols}{2}
\section{Introduction}
\hspace{0.5cm}Mobile robots are increasingly considered in applications in hostile or hazardous environments where humans cannot perform some tasks due to safety issues such as high radiation levels or because of challenges in the environments \cite{Nagatani2011,Murphy2009}. Furthermore, there exist a special category called rescue robots, which are meant to help the rescuers during disaster situation and conditions where human may not be fully able to perform the tasks due to harmful or hazardous conditions. Robots failing in such environments brings unfortunate situations such as the robot getting abandoned in the site \cite{Nagatani2011}.

\hspace{0.5cm}  In this article, some possible failure scenarios based on the experiences learned with the robots at CERN as well as from other researchers meet-up during the IEEE SSRR summer school 2012 are presented. The experiences are based on (at least) the following robots: TIM \cite{Kershaw2007} - a robotic train used for remote inspection and radiation survey for the Large Hadron Collider (LHC), Telemax - an Explosive Ordinance Disposal (EOD) robot used for remote inspection at CERN, and a KUKA youBot mobile platform used for research in robotic and manipulation applications. 

\hspace{0.5cm}Limited suggestions for design considerations aiming to reduce the chances of failures are also discussed from a simplistic perspective. For a detailed analysis on failure scenarios (how and when a mobile robot could fail), refer \cite{Carlson2005}. Additionally, observing rescue robot design standards (in progress) from NIST and ASTM \cite{Murphy2010} could also be beneficial in avoiding the failure scenarios.








\section{Hardware and software failures}
It should be considered that mechanical or electronic component failures may occur at any time during the usage of a mobile robot. We explain in this section several cases of failure or unfortunate scenarios and explain a possible pointers to avoid these problems. 
\begin{enumerate}[leftmargin=*]
\item \textit{Chronic:} When a robotic device is left switched on and was not used for a long time, they might have a tendency to become mobile due to friction loss or inertia effects. (E.g. A forklift could lose its pressure on the fork carrying a heavy load for quite a while). Therefore it should be made sure that the robot is immobile (by using a normally closed mechanical brake or the motion power) when the robot is static.
\item \textit{Calibration:} In the TIM robot \cite{Kershaw2007}, we observed that the robot consumed twice as much power when moving forward relative to backward movement. This was later discovered to be caused by a motor failure. The lesson here is that, before using a robot, it should be calibrated or verified for proper functioning of movements in all directions. This also applies to other important functional elements for e.g. sensing.
\item \textit{Thermal:} There are high chances of minor failures in robots due to thermal influences. For instance, once when we were about to give a specific demonstration with the youBot to a large audience, its onboard computer got shutdown due to overheating in a warm conference room which was unexpected. This bad experience could have been avoided if we had designed for proper heat dissipation/ventilation in the robot.

\item \textit{Over-current:} Even though short circuit problems are usually negligible in most robots as they employ professional electronics components, over current and circuit related problem may occur anytime. For instance, a USB powered WiFi card consumed too much of power from the USB port (exceeded its limit) resulting in damage to the the computer motherboard and eventual failure of the robot operation. Thus ensuring proper power availability (by using current limiter or a relay) to every devices and sensors connected to the robot could help avoiding such issues. 

\item \textit{Memory:} In one of the experiences, a sudden crash in the robot's computer hard disk resulted in instant robot failure and prevented its mission of autonomous localization and mapping of the environment. This brings up the need to have (and execute) the robot programs in redundant efficient manner, use solid state drives and also include a recovery routine in the RAM memory to switch a defective hard disk containing the main routine or atleast to shut-down the robot safely.

\item \textit{Frequency:} Over-clocking of a robots computer could result in serious damages to the robot and its performance as well. It could also trigger a spark or fire (there was a reported case that a high frequency switch got fired but then the system was immediately shut-out to avoid further damage) and therefore the robot systems and functions should carefully be designed not to be overcooked or over-clocked.

\item \textit{Faults:} If the robot is used in high energy particle physics or ionizing radiation facilities, the probability that the electronics getting damaged due to single event upset or cumulative effects of radiation are high (there were several reported incidents for such cases and for instance there exist a special group at CERN to study the effects of radiation to electronics). Hence, for such applications, we emphasize the importance of having redundant fault-tolerant hardware and software modules for important functionality as well as having fail-safe algorithms on-board the robot

\item \textit{Data verification and validation:} Autonomous robots fully depend on the data obtained from its sensors or sent to its actuators. Data related issues might occur if algorithms are not devised properly. In one instance, the laser range scanner provided false values in improper (black painted/water surface) surfaces as expected but was not accounted in the algorithm. This data was used for autonomous path planning and resulted in undesired actions/behaviors. Therefore, sensor data should always be validated for proper values even if one use a safety sensing device.

In another incident, a faulty Joystick provided error (or fault) value to the algorithm which coincided with the reverse motor actuation value, and thus resulted in robot behaving strangely by moving in the opposite of the intended direction when the joystick abruptly failed (may be due to its low battery level). A lesson here is that the range of sensor error values and the actuators input values should be strictly verified for non-coincidences, properly tuned and essentially utilized. 

\end{enumerate}



\section{Energy and Communication}

\hspace{0.5cm}Managing a mobile robot's energy and communication (wired or wireless) system form the key priority in predicting and avoiding most of the failure situations. We have observed several energy and communication failure situations of the TIM robot either when the robot ran out of energy or when it was not possible to communicate to the ground staff operating the robot. Thus, ensuring the energy and communication requirement for a robotic mission can be fulfilled before starting the mission is highly important. 

\hspace{0.5cm}A basic energy management system consists of an energy autonomy prediction system (estimating the time and distance autonomy) will be helpful in making sure that the robot meets energy expectations. This can be done by simulating the energy consumption behavior before a mission or online prediction of energy autonomy during a mission (e.g. \cite{Parasuraman2014ISMS}). 

\hspace{0.5cm}Similarly, a (wireless) communication autonomy prediction system could be employed to estimate achievable distance (range), message latency, data transfer rate, etc in a given environment. This can be included in the mission planning stage or in offline mode (not during the mission) as exemplified in \cite{Parasuraman2013IJARS} where offline prediction of wireless capabilities was examined. For example, we faced several challenges in wireless communication with robots (such as low operational ranges, frequent disconnections, abrupt quality changes, etc.) at CERN underground tunnels. This is because the nature of the environment itself is challenging for radio signal propagation due to the effects of reflections from large metallic objects or deep multipath fading. Therefore, for a successful robotic operation, monitoring of the connectivity (connection quality) and deciding what action to take in the event of loss of communication is certainly required if cases such as loss of robots in high radiation zones (e.g. abandoned Quince robot in Fukushima nuclear reactor building \cite{Nagatani2011}) should not be repeated in the future. 


\hspace{0.5cm}Finally, a vital consideration for robot practitioners is to develop/implement a functionality in robots to alert/inform the operator before the robot running out of energy or loosing wireless connectivity. This could also be extended to include situations such as when the remaining energy in robot is not enough to get the robot back to its home station or when the predicted communication capability/reach in the robot's path is not adequate. 




\bibliographystyle{IEEEtran}
\bibliography{ISR2014}

\end{multicols}
\end{document}